\documentclass[11pt,a4paper]{article}
\usepackage[nohyperref]{eacl2021}
\usepackage{times}
\usepackage{latexsym}
\usepackage[hyphens]{url}

\usepackage{microtype}
\usepackage{pgfplots}
\pgfplotsset{compat=1.14}
\usepackage{graphicx}
\usepackage{xspace}
\usepackage{pifont}
\usepackage{amsmath}
\usepackage{amssymb}
\usepackage{textcomp}
\usepackage{hyperref}
\usepackage{pifont}
\hypersetup{
    colorlinks=false
}
\usepackage{graphicx}
\usepackage{microtype}
\usepackage{comment}
\usepackage{multirow}
\usepackage{enumitem}
\usepackage{xurl}

\newcommand{\mlp}{\textsc{mlp}\xspace}
\newcommand{\cnn}{\textsc{cnn}\xspace}
\newcommand{\lstm}{\textsc{lstm}\xspace}

\newcommand{\eurlex}{\textsc{eurlex}\xspace}
\newcommand{\eu}{\textsc{eu}\xspace}
\newcommand{\uk}{\textsc{uk}\xspace}
\newcommand{\eurovoc}{\textsc{eurovoc}\xspace}

\newcommand{\ir}{\textsc{ir}\xspace}

\newcommand{\wvcent}{\textsc{w2v-cent}\xspace}
\newcommand{\drmm}{\textsc{drmm}\xspace}
\newcommand{\bm}{\textsc{bm}$_{25}$\xspace}
\newcommand{\pacrr}{\textsc{pacrr}\xspace}

\newcommand{\euuk}{\textsc{eu2uk}\xspace}
\newcommand{\ukeu}{\textsc{uk2eu}\xspace}
\newcommand{\cellar}{\textsc{cellar}\xspace}

\newcommand{\doctodoc}{\textsc{doc2doc}\xspace}
\newcommand{\bert}{\textsc{bert}\xspace}
\newcommand{\bart}{\textsc{bart}\xspace}
\newcommand{\berteu}{\textsc{legal-bert}\xspace}
\newcommand{\bertlmtc}{\textsc{c-bert}\xspace}

\newcommand{\sentbert}{\textsc{s-bert}\xspace}

\newcommand{\oracle}{\textsc{oracle}\xspace}
\newcommand{\cls}{\texttt{[cls]}\xspace}
\newcommand{\ensemble}{\textsc{ensemble}\xspace}
\newcommand{\coliee}{\textsc{coliee}\xspace}
\newcommand{\wv}{\textsc{word2vec}\xspace}

\newcommand{\nli}{\textsc{nli}\xspace}
\newcommand{\regir}{\textsc{reg-ir}\xspace}

\makeatletter
\newcommand*{\barfix}[2][.175ex]{%
  \mathpalette{\@barfix{#1}}{#2}%
}
\newcommand*{\@barfix}[3]{%
  \vbox{%
    \kern#1\relax
    \hbox{$#2#3\m@th$}%
  }%
}
\makeatother

\aclfinalcopy

\title{Regulatory Compliance through Doc2Doc Information Retrieval:\\ A case study in EU/UK legislation where text similarity has limitations}

\author{Ilias Chalkidis$^{\;\dagger\;\ddagger}$\\
{\normalsize\texttt{Ilias.Chalkidis@ey.com}}\\
{\normalsize\texttt{ichalkidis@iit.demokritos.gr}}\\\And
Manos Fergadiotis$^{\;\dagger\;\ddagger}$\\
{\normalsize\texttt{Fergadiotis.Manos@ey.com}}\\
{\normalsize\texttt{mfergadiotis@iit.demokritos.gr}}\\\AND
Nikolaos Manginas$^{\;\dagger}$\\
{\normalsize\texttt{Nikolaos.Manginas@ey.com}}\\
{\normalsize\texttt{nmanginas@iit.demokritos.gr}}\\\And
Eva Katakalou$^{\;\flat\;}$\thanks{$\;\;$The contribution of Ms. Eva Katakalou was restricted to the creation and the validation of the datasets as well as to the authoring of the corresponding parts of the manuscript.}\\
{\normalsize\texttt{e.katakalou@panteion.gr}}\\\AND
Prodromos Malakasiotis$^{\;\dagger\;\ddagger}$\\
{\normalsize\texttt{Prodromos.Malakasiotis@ey.com}}\\
{\normalsize\texttt{pmalakasiotis@iit.demokritos.gr}}\\\AND
{\mdseries$^{\dagger\;}$EY AI Centre of Excellence in Document Intelligence, NCSR ``Demokritos''}\\
$^{\ddagger\;}$Department of Informatics, Athens University of Economics and Business\\ 
$^{\flat\;}$Department of International, European and Area Studies, Panteion University\\
}

\date{}

\begin{document}
\maketitle
\begin{abstract}
Major scandals in corporate history have urged the need for \emph{regulatory compliance}, where organizations need to ensure that their controls (processes) comply with relevant laws, regulations, and policies. However, keeping track of the constantly changing legislation is difficult, thus organizations are increasingly adopting Regulatory Technology (RegTech) to facilitate the process. To this end, we introduce \emph{regulatory information retrieval} (\regir), an application of \emph{document-to-document information retrieval} (\doctodoc \ir), where the query is an entire document making the task more challenging than traditional \ir where the queries are short. Furthermore, we compile and release two datasets based on the relationships between \eu directives and \uk legislation. We experiment on these datasets using a typical two-step pipeline approach comprising a pre-fetcher and a neural re-ranker. Experimenting with various pre-fetchers from \bm to $k$ nearest neighbors over representations from several \bert models, we show that fine-tuning a \bert model on an in-domain classification task produces the best representations for \ir. We also show that neural re-rankers under-perform due to \textit{contradicting} supervision, i.e., similar query-document pairs with opposite labels. Thus, they are biased towards the pre-fetcher's score. Interestingly, applying a date filter further improves the performance, showcasing the importance of the time dimension.
\end{abstract}

\begin{figure}[!ht]
    \centering
    \includegraphics[width=\columnwidth]{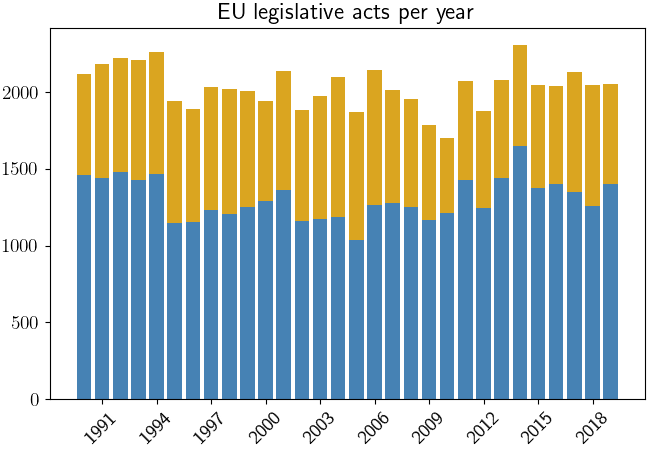}
    \caption{Number of legislative acts issued by the \eu per year. The gold color of the bars indicates how many of the published acts are amendments to older ones.}
    \label{fig:yearly_acts}
    \vspace{-2mm}
\end{figure}

\section{Introduction}

Major scandals in corporate history, from Enron to Tyco International, Olympus, and Tesco,\footnote{\url{www.theguardian.com/business/2015/jul/21/the-worlds-biggest-accounting-scandals-toshiba-enron-olympus}} have led to the emergence of stricter regulatory mandates and highlighted the need for \emph{regulatory compliance} where organizations need to ensure that they comply with relevant laws, regulations, and policies \cite{lin2016compliance}. However, keeping track of the constantly changing legislation (Figure~\ref{fig:yearly_acts}) is hard, thus organizations are increasingly adopting Regulatory Technology (RegTech) to facilitate the process.

Typically, a compliance regimen includes three distinct but related types of measures, \emph{corrective}, \emph{detective}, and \emph{preventive} \cite{Sadiq2015}. Corrective measures are usually undertaken when new regulations are introduced to update existing controls. Detective measures, ensure ``after-the-fact'' compliance, i.e., following a procedure, a manual or automated check is carried out, to ensure that every step of the procedure complied with the corresponding regulations. Finally, preventive measures ensure compliance ``by design'', i.e., during the creation of new controls. All types of measures include an underlying information retrieval (\ir) task, where laws need to be retrieved given a control or vice versa. We identify two use cases:
\begin{enumerate}
    \itemsep0em
    \item \emph{Given a new law retrieve all the controls of the organization affected by this law}. The organization can then apply corrective measures to ensure compliance for these controls.
    
    \item \emph{Given a control retrieve all relevant laws the control should comply with}. This is useful for ensuring compliance after a procedure has been carried out (detective measures) or when creating new controls (preventive measures). 
\end{enumerate}

\noindent\emph{Regulatory information retrieval} (\regir), similarly to other applications of \emph{document-to-document} (\doctodoc) \ir, is much more challenging than traditional \ir where the query typically contains a few informative words and the documents are relatively small (Table~\ref{tab:ir_datasets_stats}). In \doctodoc \ir the query is a long document (e.g., a regulation) containing thousands of words, most of which are uninformative. Consequently, matching the query with other long documents where the informative words are also sparse, becomes extremely difficult.

\begin{figure}[t]
    \centering
    \includegraphics[width=\columnwidth]{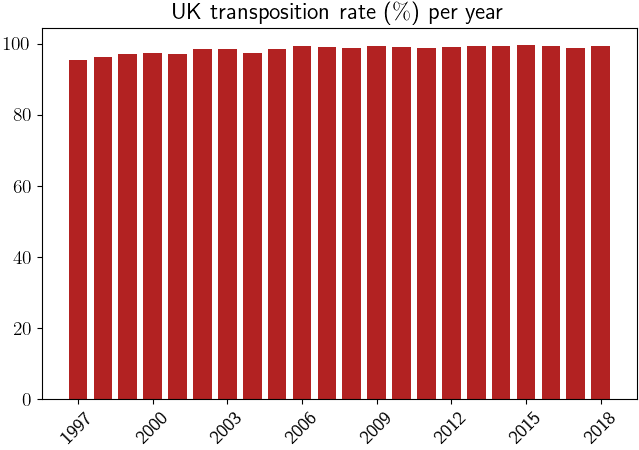}
    \caption{The percentage of \eu directives transposed by \uk legislation per year. Over 98\% of the published \eu directives have been transposed.}
    \label{fig:transpositions}
\end{figure}

Although legislation is available, organizations' controls are strictly private and very hard to obtain. Fortunately, the European Union (\eu) has a legislation scheme analogous to regulatory compliance for organizations. According to the Treaty on the Functioning of the European Union (\textsc{tfeu}),\footnote{Articles 291 (1) and 288 paragraph 3.} all published \eu \emph{directives} must take effect at the national level. Thus, all \eu member states must adopt a law to transpose a newly issued directive within the period set by the directive (typically 2 years). Notably, the United Kingdom (\uk) having a high compliance level with the \eu (Figure~\ref{fig:transpositions}),\footnote{Data for Figures~\ref{fig:yearly_acts} and \ref{fig:transpositions} obtained from \url{ec.europa.eu/internal_market/scoreboard/performance_by_governance_tool/eu_pilot}.} is a good test-bed for \regir. Thus we compile and release two datasets for \regir, \euuk and \ukeu, containing \eu directives and \uk regulations, which can serve both as queries and documents under the ground truth assumption that a \uk law is relevant to the \eu directives it transposes and vice versa.

\begin{table}[ht!]
    \centering
    \footnotesize{
    \resizebox{\columnwidth}{!}{
    \begin{tabular}{l|c|c|c}
    \hline\hline
    Dataset                   & Domain & $\tilde{q}$ & $\tilde{d}$ \\\hline
    \multicolumn{4}{c}{\emph{IR datasets in the literature}} \\ \hline
    \textsc{trec robust} \citep{trecrobust}       & News                       & 3 / 14 & 254  \\
    \textsc{bioasq} \citep{Tsatsaronis2015}           & Biomedical                 & 9      & 197 \\ \hline
    \multicolumn{4}{c}{\emph{IR datasets with verbose queries}} \\ \hline
    \textsc{gov2} \citep{Clarke2004TrecTerabyte}       & Web                & 11 / 57  & 682   \\
    \textsc{wt10g} \citep{wt10_dataset}            & Web                & 11 / 35  & 457  \\ \hline
    \multicolumn{4}{c}{\emph{Regulatory Compliance datasets}} \\ \hline
    \euuk  (ours)             & Law                & 2,642  & 1,849   \\
    \ukeu  (ours)            & Law                & 1,849  & 2,642  \\ \hline\hline
    \end{tabular}
    }
    }
    \caption{Statistics for query and document length for IR datasets used in literature.}
    \label{tab:ir_datasets_stats}
\end{table}

\begin{table*}[t!]
    \centering
    \resizebox{\textwidth}{!}{
    \begin{tabular}{c|c|l}
    \hline\hline
    \multicolumn{3}{p{8in}}{\textbf{Query}: DIRECTIVE 2006/66/EC OF THE EUROPEAN PARLIAMENT AND OF THE COUNCIL of 6 September 2006 on batteries and accumulators and waste batteries and accumulators and repealing Directive 91/157/EEC}\\\hline
    \bm rank             & Relevant     & \multicolumn{1}{|c}{Document title}\\\hline
    1                    & No           & The Batteries and Accumulators (Placing on the Market) (Amendment) Regulations 2012\\
    2                    & No           & The Batteries and Accumulators (Containing Dangerous Substances) (Amendment) Regulations 2000\\
    3                    & No           & The Batteries and Accumulators (Placing on the Market) (Amendment) Regulations 2015\\
    4                    & No           & The Batteries and Accumulators (Containing Dangerous Substances) Regulations 1994\\
    5                    & No           & The Waste Batteries and Accumulators (Amendment) Regulations 2015\\
    6                    & Yes          & The Waste Batteries and Accumulators Regulations 2009\\
    12                   & Yes          & The Batteries and Accumulators (Placing on the Market) Regulations 2008\\\hline\hline
    \end{tabular}
    }
    \caption{Example from the \euuk dataset where the retrieved \uk laws are ranked by \bm. The top-5 documents seem similar to the query but are not relevant. Documents ranked 1st, 3rd, and 5th are amendments of the relevant documents, i.e., \uk laws that transpose the query.}
    \label{tab:examples}
\end{table*}

Since \regir is a new task, our starting point is the two-step pipeline approach followed by most modern neural information retrieval systems \cite{guo2016,hui2017pacrr,mcdonald2018}. First, a conventional \ir system (\emph{pre-fetcher}) retrieves the $k$ most prominent documents. Then a neural model attempts to rank relevant documents higher than irrelevant ones. In most approaches, the pre-fetcher is based on Okapi \bm \cite{robertson1995}, a bag-of-words scoring function that does not consider possible synonyms or contextual information. To overcome the first limitation, we follow \newcite{brokos2016} who employed $k$ nearest neighbors over $\mathrm{tf}$-$\mathrm{idf}$ weighted centroids of word embeddings, without however improving the results, probably because the centroids are noisy considering many uninformative words. Furthermore, we employ \bert \cite{devlin2019} to extract contextualized representations for queries and documents but again the results are worse than \bm. We also experiment with \sentbert \cite{reimers2019sentence} and \berteu \citep{chalkidis-etal-2020-legal}, a model specialized in the legal domain. Both models perform better than \bert but are still worse than or comparable to \bm. The inability of \bert{-based} models motivated us to find an auxiliary task that will result in better representations for \regir. Following \newcite{chalkidis2019}, we fine-tune \bert to predict \eurovoc concepts that describe the core subjects of each text. As expected this model (\bertlmtc) is the best pre-fetcher by a large margin in \euuk, while being comparable to \bm in \ukeu. 
To summarize, our contributions are:
\begin{enumerate}[label=(\alph*)]
    \itemsep0em
    \item We introduce \regir, an application of \doctodoc \ir, which is a new family of \ir tasks, where both queries and documents are long typically containing thousands of words.
    \item We compile and release the two first publicly available datasets, \euuk and \ukeu, suitable for \regir and \doctodoc \ir in general.\footnote{The datasets are available at \url{https://archive.org/details/eacl2021_regir_datasets}.}
    \item We show that fine-tuning \bert on an in-domain classification task produces the best document representations with respect to \ir and improves pre-fetching results.
\end{enumerate}

\begin{table*}[ht!]
    \centering
    \footnotesize{
    \begin{tabular}{c|c|c|c|c|c|c|c}
    \hline\hline
    \multirow{2}{*}{Dataset} & Documents & \multicolumn{2}{c|}{Train} & \multicolumn{2}{c|}{Development} & \multicolumn{2}{c}{Test}\\\cline{3-8}
                             & in pool   & Queries & Avg. relevant    & Queries & Avg. relevant          & Queries & Avg. relevant\\\hline
    \euuk                    & 52,515    & 1,400   & 1.79             & 300     & 2.09                   & 300     & 1.74\\
    \ukeu                    &  3,930    & 1,500   & 1.90             & 300     & 1.46                   & 300     & 1.29\\\hline\hline
    \end{tabular}
    }
    \caption{Detailed statistics for \euuk and \ukeu. Both datasets have relatively small number of relevant documents while \euuk has also large pool which may impose extra difficulties in the retrieval.}
    \label{tab:datasets_stats}
\end{table*}

\section{Datasets curation}

\subsection{Data sources}

\noindent\textbf{EU/UK Legislation:} We have downloaded approx. 56K pieces of \eu legislation (approx. 3.9K directives), from the \eurlex portal.\footnote{\url{eur-lex.europa.eu}} \eu laws are 2,642 words long on average and are structured in three major parts: the \emph{title} (Table~\ref{tab:examples}, query), the \emph{recitals} consisting of references in the legal background of the act, and the \emph{main body}. We have also downloaded approx. 52K \uk laws, publicly available from the official \uk legislation portal.\footnote{\url{legislation.gov.uk}} \uk laws are 1,849 words long on average and contain the \emph{title} (Table~\ref{tab:examples}, document title) and the \emph{main body}.
\vspace{1.5mm}

\noindent\textbf{Transpositions:} We have retrieved all transposition relations (approx. 3.7K) between \eu directives and \uk laws from the \cellar database. \cellar only provides the mapping between the \cellar ids of \eu directives and the title of each \uk law. Therefore we aligned the \cellar ids with the official \uk ids based on the law title.\footnote{See Appendix~\ref{sec:technical} for details on the dataset curation.} One or more \uk laws may transpose one or more \eu directives.

\subsection{Datasets compilation}
\label{sec:data_compilation}

Let $\mathcal{E}$, $\mathcal{U}$ be the sets of \eu directives and \uk laws, respectively. We define \regir as the task where the query $q$ is a document, e.g, an \eu directive, and the objective is to retrieve a set of relevant documents, $\mathcal{R}_q$, from the pool of all available documents, e.g., all \uk laws. We create two datasets:\vspace{1.5mm}

\noindent\textbf{\euuk:} $q\in\mathcal{E}$, $\mathcal{R}_q = \{r_i\!: r_i \in \mathcal{U}, r_i\!\xrightarrow{\text{transposes}}\!q\}$.\vspace{1.5mm}

\noindent\textbf{\ukeu:} $q\in\mathcal{U}$, $\mathcal{R}_q = \{r_i\!: r_i\in \mathcal{E}, q\!\xrightarrow{\text{transposes}}\!r_i\}$.\vspace{1.5mm}

Table~\ref{tab:datasets_stats} shows the statistics for the two datasets, which are split in three parts, \emph{train}, \emph{development}, and \emph{test}, retaining a chronological order for the queries. \euuk has a much larger pool of available documents than \ukeu (52.5K vs.\ 3.9K) which may impose an extra difficulty during retrieval. More importantly, the average number of relevant documents per query is small (at most 2) for both datasets, as our ground truth assumption is strict, i.e., relevant documents are those linked to the query with a transposition relation. Also, \eu legislation is frequently amended (Figure~\ref{fig:yearly_acts}) which also imposes difficulty in the retrieval task. Let $d_1\in\mathcal{E}$ be a directive transposed by $u_1\in\mathcal{U}$ and $d_2\in\mathcal{E}$ be a directive amending $d_1$. The \uk must adopt a law, $u_2$, to transpose $d_2$. Both $d_2$ and $u_2$ cover similar concepts to those of $d_1$ ($d_2$ is an amendment and $u_2$ must comply with $d_2$), but, strictly speaking $u_2$ is relevant only to $d_2$. Table~\ref{tab:examples} shows an example from \euuk, where the top-5 documents seem very similar to the query but are not considered relevant. Note that the documents ranked 1st, 3rd and 5th, are amendments of the relevant documents.

\section{IR pipelines}

Modern neural \ir systems usually follow a two-step pipeline approach. First, a conventional \ir system (\emph{pre-fetcher}) retrieves the top-$\mathrm{k}$ most prominent documents aiming to maximize its recall. Then a neural model attempts to re-rank the documents by scoring relevant higher than irrelevant ones. While this configuration is widely adopted in literature, the re-ranking step could be omitted provided an effective pre-fetching mechanism, i.e., the pre-fetcher will act as an end-to-end \ir system.

\subsection{Document pre-fetching}
\label{sec:prefetching}

\textbf{Okapi \bm} \cite{robertson1995} is a bag-of-words scoring function estimating the relevance of a document $d$ to a query $q$, based on the query terms appearing in $d$, regardless their proximity within $d$:
\begin{equation}
\sum\limits_{i=1}^{n}\mathrm{idf}(q_{i})\cdot\frac{\mathrm{tf}(q_{i},d)\cdot(k_{1}+1)}{\mathrm{tf}(q_{i},d)+k_{1}\cdot\left(1-b+b\cdot\frac{L}{\barfix{\bar L}}\right)}
\end{equation}

\noindent where $q_i$ is the $i$-th query term, with $\mathrm{idf}(q_i)$ inverse document frequency and $\mathrm{tf}(q_i, d)$ term frequency. $L$ is the length of $d$ in words, $\bar{L}$ is the average length of the documents in the collection, $k_{1}$ is a parameter that favors high $\mathrm{tf}$ scores and $b$ is a parameter penalizing long documents.\footnote{We use \emph{elastic}, a widely used \ir engine with the \bm scoring function. See \url{www.elastic.co/}.}\vspace{1.5mm}

\noindent\textbf{\wvcent}: Following \newcite{brokos2016}, we represent query/document terms with pre-trained embeddings. For each query/document we calculate the $\mathrm{tf}$-$\mathrm{idf}$ weighted centroid of its embeddings:
\begin{equation}
    \mathrm{cent}(t) = \frac{\sum_{i=1}^{l}\boldsymbol{x}_i\cdot\mathrm{tf}(x_i,t)\cdot\mathrm{idf}(x_i)}{\sum_{i=1}^{l}\mathrm{tf}(x_i,t)\cdot\mathrm{idf}(x_i)}
\end{equation}

\noindent where $t$ is a text (query or document) and $x_i$ is the $i$-th text term with embedding $\boldsymbol{x}_i$. The documents are ranked, with respect to the query, by a $\mathrm{k}$ nearest neighbours ($\mathrm{kNN}$) algorithm with cosine distance:
\begin{equation}
    \mathrm{cos_d}(q,d) = 1 - \frac{\mathrm{cent}(q)\boldsymbol{\cdot}\mathrm{cent}(d)}{\|\mathrm{cent}(q)\|\cdot\|\mathrm{cent}(d)\|}
\end{equation}

\noindent\textbf{\bert}, similarly to \wvcent, relies in pre-trained representations which now are extracted from \bert, thus being context-aware. A text can be represented by its \cls token or by the centroid of its token embeddings. In the latter case the embeddings can be extracted from any of the 12 layers of \bert.\footnote{\bert is not fine-tuned during this process.} Note that the texts in our datasets do not entirely fit in \bert. We thus split them into $c$ chunks (2 to 3 per text) and pass each chunk through \bert to obtain a list of token embeddings per layer (i.e, the concatenation of $c$ token embeddings lists) or $c$ \cls tokens. The final representation is either the centroid of the token embeddings or the centroid of the \cls tokens.\vspace{1.5mm}

\noindent\textbf{\sentbert} \cite{reimers2019sentence} is a \bert model fine-tuned for \nli. According to the authors, training \sentbert for \nli results in better representations than \bert for tasks involving text comparison, like \ir. We use the same setting as in \bert.\vspace{1.5mm}

\noindent\textbf{\berteu}: Our datasets come from the legal domain which has distinct characteristics compared to generic corpora, such as specialized vocabulary, particularly formal syntax, semantics based on extensive domain-specific knowledge, etc., to the extent that legal language is often classified as a `sublanguage' \cite{Tiersma1999,Williams2007,Haigh2018}. \bert and \sentbert were trained on generic corpora and may fail to capture the nuances of legal language. Thus we used a \bert model further pre-trained on \eu legislation \cite{chalkidis-etal-2020-legal}, dubbed here \berteu, in a similar fashion.\vspace{1.5mm}

\noindent\textbf{\bertlmtc}: \eu laws are annotated with \eurovoc concepts covering the core subjects of \eu legislation (e.g., environment, trade, etc.). Our intuition is that a \uk law transposing an \eu directive will most probably cover the same subjects. Thus we expect that a \bert model, fine-tuned to predict \eurovoc concepts, will learn rich representations describing these concepts which may be useful for pre-fetching. We fine-tune \bert following \newcite{chalkidis2019}\footnote{We use all \eu laws excluding \eu directives that exist in our development and test sets.} and use the resulting model to extract query and document representations similarly to the previous \bert{-based} methods.\vspace{2mm}

\noindent\textbf{\ensemble} is simply a combination of our best two pre-fetchers, \bertlmtc and \bm:
\begin{equation}
    \textsc{ens}(q, d) = \alpha\cdot\textsc{cb}(q, d) + (1-\alpha)\cdot\textsc{bm}_{25}(q,d)
\end{equation}

\noindent where \textsc{cb} is the score of \bertlmtc and $\alpha$ is tuned on development data and the scores of the pre-fetchers are normalized in $[0, 1]$.

\subsection{Document re-ranking}
\label{sec:reranking}

Modern neural re-rankers operate on pairs of the form $(q, d)$ to produce a relevance score, $\mathrm{rel}(q, d)$, for a document $d$ with respect to a query $q$. Note, however, that the main objective is to rank relevant documents higher than irrelevant. Thus, during training the loss is calculated as:
\begin{equation}
\mathcal{L} = \max(0, 1 - \mathrm{rel}(q, d^{+}) + \mathrm{rel}(q, d^{-}))
\end{equation}

\noindent where $d^{+}$ is a relevant document and $d^{-}$ is an irrelevant document. We have experimented with several neural re-ranking methods each having a function that produces a relevance score $s_r$ for each of the top-$\mathrm{k}$ documents returned by the best pre-fetcher. The final relevance score of a document is calculated as: $\mathrm{rel}(q, d)=w_r\cdot s_r+w_p\cdot s_p$, where $s_p$ is the normalized score of the pre-fetcher and $w_s$, $w_p$ are learned during training. 

Given the concerns on the strictness of the ground truth assumption raised in Section~\ref{sec:data_compilation}, we hypothesize that re-rankers will eventually over-utilize the pre-fetcher score, $s_p$, when calculating document relevance, $\mathrm{rel}(q, d)$. As shown in Table~\ref{tab:examples}, in many cases both relevant and irrelevant documents may have high similarity with the query. This in turn may confuse and therefore degenerate the re-ranker's term matching mechanism, i.e., \mlp{s} or \cnn{s} over term similarity matrices.\vspace{1.5mm}

\noindent\textbf{\drmm} \cite{guo2016} uses pre-trained word embeddings to represent query and document terms. A histogram captures the cosine similarities of a query term, $q_i$, with all the terms of a particular document. Then an \mlp consumes the histograms to produce a document-aware score for each $q_i$, which is weighted by a gating mechanism assessing the importance of $q_i$. The sum of the weighted scores is the relevance score of the document. A caveat of \drmm is that it completely ignores the context of the terms which could be of particular importance in our datasets where texts are long.\vspace{1.5mm}

\noindent\textbf{\pacrr} \cite{hui2017pacrr} represents query and document terms with pre-trained embeddings and calculates a matrix $S$ containing the cosine similarities of all query-document term pairs. A row-wise $k$-$\max$ pooling operation on $S$ keeps the highest similarities per query term (matrix $S_k$). Then, wide convolutions of different kernel (filter) sizes ($n\times n$) with multiple filters per size are applied on $S$. Each filter of size $n\times n$ attempts to capture $n$-gram similarities between queries and documents. A $\max$-pooling operation keeps the strongest signals across filters and a row-wise $k$-$\max$ pooling keeps the strongest signals per query $n$-gram, resulting in the matrix $S_{n,k}$. Subsequently, a row-wise concatenation of $S_k$ with all $S_{n,k}$ matrices (for different values of $n$) is performed and a column containing the $\mathrm{softmax}$-normalized $\mathrm{idf}$ scores of the query terms is concatenated to the resulting matrix ($S_\mathrm{sim}$). In effect, each row of the matrix contains different $n$-gram based similarity views of the corresponding query term, $q_i$, along with an $\mathrm{idf}$-based importance score. The relevance score is produced as the last hidden state of an \lstm with one hidden unit, which consumes the rows of $S_\mathrm{sim}$. \pacrr tries to take into account the context of the query and document terms using $n$-grams but this context sensitivity is weak and we do not expect much benefits in our datasets which contain long texts.\vspace{1.5mm}

\begin{figure*}[!ht]
    \centering
    \includegraphics[width=0.9\textwidth]{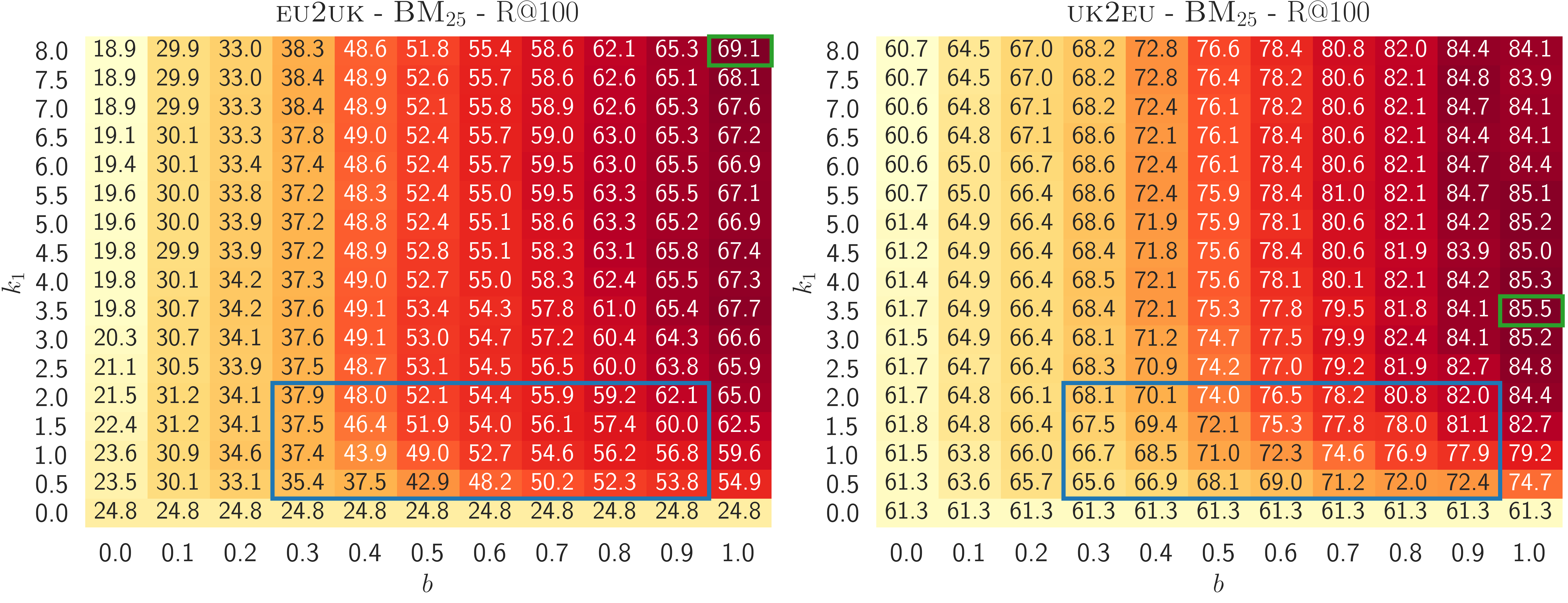}
    \caption{Heatmaps showing $\mathrm{R@100}$ for different values of $k_1$ and $b$ on \euuk (left) and \ukeu (right). The selected optimal values (green boxes) are outside the proposed ranges in the literature (blue boxes).}
    \label{fig:heatmaps}
\end{figure*}

\noindent\textbf{\bert{-based} re-rankers}: Recent work tries to exploit \bert to improve re-ranking. Following \newcite{cedr2019}, we use \drmm and \pacrr on top of contextualized \bert embeddings derived from \bert. Based on the results of Figure~\ref{fig:bert_heatbars}, we use \bertlmtc as the most promising \bert model. We call these two models \bertlmtc-\drmm and \bertlmtc-\pacrr. We also experiment with two settings depending on whether \bertlmtc weights are updated (\emph{tuned}) or not (\emph{frozen}) during training.\vspace{1.5mm}

\section{Experimental setup}

\subsection{Pre-trained resources}

As several methods rely on word embeddings, we trained a new \wv model \cite{Mikolov2013} in both corpora (\eu and \uk legislation) to better accommodate legal language. Preliminary experiments showed that domain-specific embeddings perform better than generic 200-dimensional GloVe embeddings \cite{pennington2014glove} in development data (\euuk: 66.5 vs. 59.3  at $\mathrm{R@100}$ and \ukeu: 72.6 vs. 69.8 at $\mathrm{R@100}$).\footnote{See also the discussion for legal language in Section~\ref{sec:prefetching}.}

All \bert (pre-fetching) encoders and \bert{-based} re-rankers use the \textsc{-base} version, i.e., 12 layers, 768 hidden units and 12 attention heads, similar to the one of \citet{devlin2019}.\footnote{See Appendix~\ref{sec:bertmodels} for more details.}

\subsection{Pre-processing - document denoising}
One of the major challenges in \doctodoc \ir, as opposed to traditional \ir, is the length of the queries and the documents which may induce noise (many uninformative words) during retrieval. Thus we applied several filters (stop-word, punctuation and digits elimination) on both queries and documents and reduced their length by approx. 55\% (778 words for \uk laws and 1,222 words for \eu directives on average). Further on, we filtered both queries and documents by eliminating words with $\mathrm{idf}$ score less than the average $\mathrm{idf}$ score of the stop-words. Our intuition is that words (e.g., regulation, \eu, law, etc.) with such a small $\mathrm{idf}$ score are uninformative. Still, the texts are much longer (387 words for \uk laws and 631 words for \eu directives on average) than the queries used in traditional \ir (Table~\ref{tab:ir_datasets_stats}). As an alternative to drastically decrease the query size, we experimented with using only the title of a legislative act as a query but the results were worse, i.e., approx. 5-20\% lower $\mathrm{R@100}$ on average across datasets, indicating that the full-text is more informative, although the information is sparse. Hence, we only consider the full-text, including the title, for the rest of the experiments.

\subsection{Evaluation measures}
Pre-fetching aims to bring all the relevant documents in the top-$\mathrm{k}$, thus we report $\mathrm{R@k}$. We observe that for $\mathrm{k>100}$ the best pre-fetchers have not significant gains in performance in development data, thus we select $\mathrm{k=100}$, as a reasonable threshold.\footnote{See Appendix~\ref{sec:appendix_transpositions} for an extended ($\mathrm{k} \in [0,2000]$) performance evaluation on pre-fetching.} For re-ranking we report $\mathrm{R@20}$, $\mathrm{nDCG@20}$ and $\mathrm{R}$-$\mathrm{Precision}$ ($\mathrm{RP}$) following the literature \cite{Manning2009}. We report the average and standard deviation across three runs considering the best set of hyper-parameters on development data for neural re-rankers.

\begin{figure*}[!ht]
    \centering
    \includegraphics[width=0.9\textwidth]{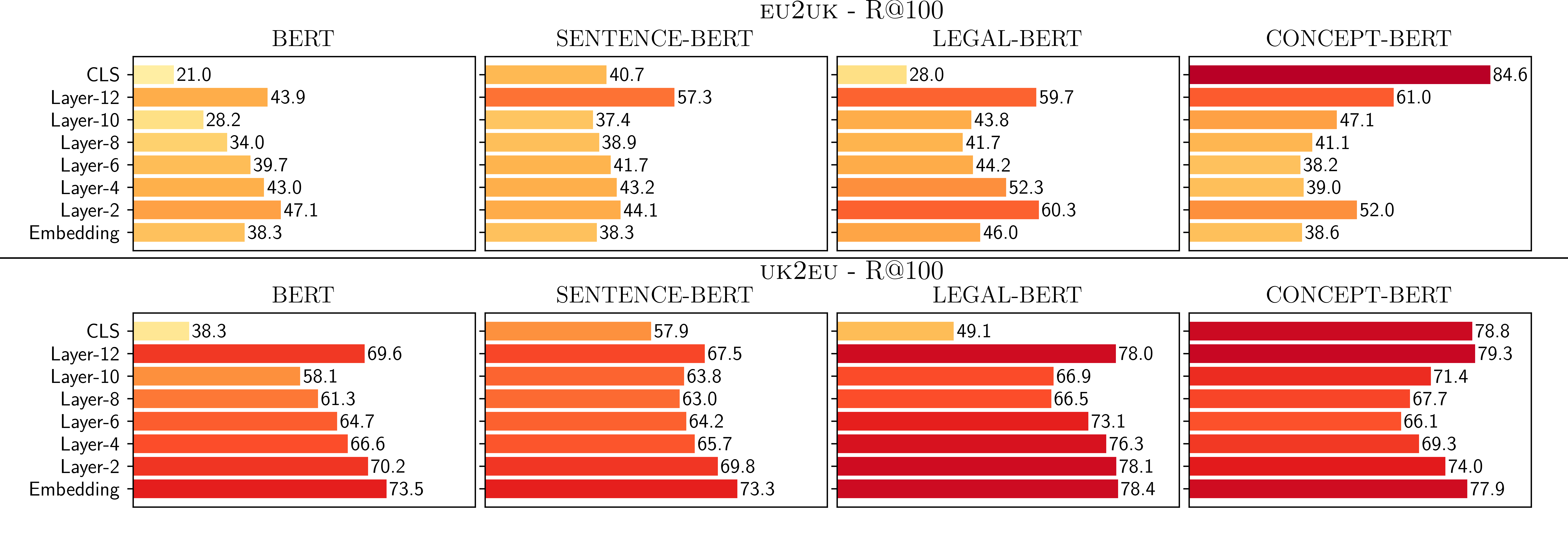}
    \caption{Heatbars showing $\mathrm{R@100}$ (on development data) for text representations extracted from different layers of the various \bert{-based} pre-fetchers we experimented with.}
    \label{fig:bert_heatbars}
\end{figure*}

\subsection{Tuning BM$_{25}$: The case of \doctodoc IR}
The effectiveness of \bm is highly dependant on properly selecting the values of $k_1$ and $b$. In traditional (ad-hoc) \ir, $k_1$ is typically evaluated in the range$[0,3]$ (usually $k_1\in[0.5, 2.0]$); $b$ needs to be in $[0, 1]$ (usually $b\in[0.3, 0.9]$) \cite{Taylor2006,Trotman2014,Lipani2015}. As a general rule of thumb \bm with $k_1$=1.2 and $b$=0.75 seems to give good results in most cases \cite{Trotman2014}. We observe that in the case of \doctodoc \ir where the queries are much longer, the optimal values are outside the proposed ranges (Figure~\ref{fig:heatmaps}). In both datasets the optimal values for $k_1$ and $b$ are relatively high, favoring terms with high $\mathrm{tf}$, while penalizing long documents. In effect \bm  uses $k_1$ and $b$ as a denoising regularizer to over-utilize highly frequent query terms normalized by document length.

\subsection{Extracting representations from BERT}
\label{sec:bert-represntation}
Recently there has been a lot of research on understanding the effectiveness of \bert{'s} different layers \cite{liu2019linguistic,hewitt2019structural,jawahar2019bert,Goldberg2019,Kovaleva2019secrets,Lin2019Sesame}. Figure~\ref{fig:bert_heatbars} shows heatbars comparing representations extracted from different layers of the various \bert{-based} pre-fetchers we experimented with.\footnote{Recall that a text can be represented by its \cls token or by the centroid of its token embeddings which can be extracted from any of the 12 layers of \bert.} \berteu and \bertlmtc which have been adapted in the legal domain perform much better than \bert and \sentbert which were trained on generic corpora. An interesting observation is that the \cls token is a powerful representation only in \bertlmtc where it was trained to predict \eurovoc concepts. Also, in \ukeu the embedding layer produces the best representations in all \bert variants except \bertlmtc, where the embedding layer achieves comparable results to the top-2 representations (\cls, Layer-12). This is an indication that the context in this dataset is not as important as in \euuk.

\subsection{Implementation details}
All neural models were implemented using the Tensorflow 2 framework. Hyper-parameters were tuned on development data, using early stopping and the Adam optimizer \cite{Kingma2015}.

\begin{table}[t!]
    \centering
    \resizebox{\columnwidth}{!}{
    \begin{tabular}{l|c|c}
    \hline\hline
         \hfil\multirow{2}{*}{Method}\hfill & \euuk & \ukeu \\
         \cline{2-3}
                                              & $\mathrm{R@100}$ & $\mathrm{R@100}$\\ \hline
         \bm \cite{robertson1995}             & 57.5             & \underline{93.7}          \\
         \wvcent \cite{brokos2016}            & 50.6             & 88.2            \\
         \bert \cite{devlin2019}              & 54.0             & 85.1            \\
         \sentbert \cite{reimers2019sentence} & 57.7             & 84.8            \\
         \berteu \citep{chalkidis-etal-2020-legal}                       & 57.6             & 90.1            \\
          \bertlmtc (ours)                    & \underline{83.8}            & \underline{92.9}            \\
          \hline
          \ensemble (\bm + \bertlmtc)         & \textbf{86.5}    & \textbf{95.0}   \\
         \hline\hline
    \end{tabular}
    }
    \caption{Pre-fetching results across test datasets.}
    \label{tab:pre-fetching}
    \vspace{-3mm}
\end{table}

\begin{table*}[ht!]
    \centering
    \resizebox{\textwidth}{!}{
    \begin{tabular}{l|cc|ccc|cc|ccc}
    \hline\hline
         \hfil\multirow{2}{*}{Method}\hfill & \multicolumn{5}{c|}{\euuk}                             & \multicolumn{5}{c}{\ukeu}\\\cline{2-11}
                                                            & $w_p$ & $w_s$ & $\mathrm{R@20}$           & $\mathrm{nDCG}@20$        & $\mathrm{RP}$            & $w_p$ & $w_s$ & $\mathrm{R@20}$           & $\mathrm{nDCG}@20$        & $\mathrm{RP}$ \\\hline
         \bm                                &   -   & -     & 45.8                      & 34.4                      & 25.5                      &   -   &  -    & 87.5                      & 66.8                      & \textbf{49.4} \\
         \bertlmtc (ours)                   &   -   & -     & 55.7                      & 37.9                      & 21.8                      &   -   &  -    & 79.7                      & 53.0                      & 33.1 \\\hline
         \ensemble (\bm + \bertlmtc)        &   -   & -     & 54.1                      & 43.1                      & 29.6                      &   -   &  -    & 88.0                      & \textbf{67.7}             & 49.3 \\
         + \drmm                            & +1.1  & -0.8  & \textbf{59.9} ($\pm$ 3.2) & 41.7 ($\pm$ 2.4)          & 24.3 ($\pm$ 2.9)          & +1.3  & -0.8  & 86.3 ($\pm$ 1.1)          & 61.6 ($\pm$ 1.1)          & 40.1 ($\pm$ 1.5)\\
         + \pacrr                           & +4.2  & +0.6  & 54.3 ($\pm$ 0.2)          & \textbf{43.3} ($\pm$ 0.2) & \textbf{30.1} ($\pm$ 0.4) & +4.0  & +0.1  & 88.0 ($\pm$ 0.0)          & \textbf{67.7} ($\pm$ 0.0)          & 49.3 ($\pm$ 0.0)\\
         + \bertlmtc-\drmm \emph{(frozen)}  & +3.3  & -1.6  & 57.9 ($\pm$ 3.4)          & 43.1 ($\pm$ 0.3)          & 27.3 ($\pm$ 2.2)          & +3.5  & -1.0  & 88.3 ($\pm$ 0.4)          & 67.3 ($\pm$ 0.6)          & 48.5 ($\pm$ 1.3)\\
         + \bertlmtc-\pacrr \emph{(frozen)} & +4.6  & +0.9  & 54.1 ($\pm$ 0.0)          & 43.1 ($\pm$ 0.0)          & 29.6 ($\pm$ 0.0)          & +2.9  & -0.9  & \textbf{89.6} ($\pm$ 0.4) & 66.5 ($\pm$ 0.5)          & 46.0 ($\pm$ 0.9)\\
         + \bertlmtc-\drmm \emph{(tuned)}   & +1.9  & -0.5  & 54.1 ($\pm$ 0.0)          & 43.1 ($\pm$ 0.0)          & 29.6 ($\pm$ 0.0)          & +1.2  & +0.5  & 88.0 ($\pm$ 0.0)          & \textbf{67.7} ($\pm$ 0.0) & 49.3 ($\pm$ 0.0)\\
         + \bertlmtc-\pacrr \emph{(tuned)}  & +1.8  & -0.6  & 54.1 ($\pm$ 0.0)          & 43.1 ($\pm$ 0.0)          & 29.6 ($\pm$ 0.0)          & +2.0  & +2.1  & 88.0 ($\pm$ 0.0)          & \textbf{67.7} ($\pm$ 0.0)          & 49.3 ($\pm$ 0.0)\\
         \hline
         + \oracle                          &   -   &   -   & 86.5                      & 87.7                      & 86.5                      &   -   &   -   & 95.0                      & 95.3                      & 95.0 \\
         \hline
         \multicolumn{11}{c}{\emph{Applying date filtering on top of predictions}} \\
         \hline
         Year range & \multicolumn{5}{c|}{$\pm5$ years} & \multicolumn{5}{c}{$\pm15$ years} \\
         \hline
        \ensemble (\bm + \bertlmtc)         &   -   &   -   & 76.6                      & 54.6                      & 37.1             &   -   &   -   & \textbf{86.2}                      & \textbf{68.2}             &\textbf{50.0} \\ 
         + \drmm (\emph{pre-filtering})     & +1.1  & -0.8  & \textbf{81.4} & \textbf{56.5} & 35.4           & +1.3  & -0.8  & 85.3 & 62.6 & 42.3 \\ 
         + \drmm  (\emph{post-filtering})   & +1.1  & -0.8  & 75.7 & 49.2 & 31.1           & +1.3  & -0.8  & 83.6 & 63.5 & 44.2 \\ 
         \hline
         + \pacrr (\emph{pre-filtering})     & +4.2  & +0.6  & 76.6 & 54.8 & \textbf{37.6}          & +4.0  & +0.1  & \textbf{86.2} & \textbf{68.2} & \textbf{50.0} \\ 
         + \pacrr  (\emph{post-filtering})   & +4.2  & +0.6  & 74.2 & 52.9 & 36.5 & +4.0  & +0.1  & 85.5 & 67.6 & 49.6 \\ 
         \hline\hline
    \end{tabular}
    }
    \caption{Re-ranking results across test datasets. The upper zone shows the results of neural re-rankers on top of the best pre-fetchers with respect to ($w_s$, $w_p$). It also reports re-ranking results of the best pre-fetchers. The lower zone reports the re-ranking results after applying temporal filtering.}
    \label{tab:re-ranking}
\end{table*}

\section{Experimental results}

\textbf{Pre-fetching:} Table~\ref{tab:pre-fetching} shows $\mathrm{R@100}$ on the test datasets for the various pre-fetchers considered. On \euuk, \bertlmtc is the best method by a large margin, followed by \sentbert and \berteu, verifying our assumption that the concept classification task is a good proxy for obtaining rich representations with respect to \ir. Both \sentbert and \berteu are better than \bert for different reasons. \berteu was adapted to the legal domain and is, therefore, able to capture the nuances of the legal language. \sentbert was trained to produce representations suitable for comparing texts with cosine similarity, a task highly related to \ir. Nonetheless, having been trained on generic corpora with small texts, it performs much worse than \bertlmtc. Interestingly, \bm is comparable to both \sentbert and \berteu despite its simplicity. As expected, combining \bertlmtc with \bm further improves the results. In \ukeu $\mathrm{R@100}$ is much higher compared to \euuk probably because of the shortest queries. Also, as discussed in Section~\ref{sec:bert-represntation}, the contextual information is not so critical in this dataset, thus we expect the context unaware \bm and \wvcent to perform well. Indeed, \bm achieves the best results followed closely by \bertlmtc and \berteu, while \wvcent outperforms \sentbert and \bert. Again the \ensemble improves the results.\vspace{1mm}

\noindent\textbf{Re-ranking:} Table~\ref{tab:re-ranking} shows the ranking results on test data for \euuk and \ukeu. We also report results for \bm, \bertlmtc, \ensemble and an \oracle, which re-ranks the top-$\mathrm{k}$ documents returned by the pre-fetcher placing all relevant documents at the top. On \euuk \ensemble performs better than the other two pre-fetchers. Interestingly, neural re-rankers fall short on improving performance and are comparable (or even identical) with \ensemble in most cases, possibly because very similar documents may be relevant or not (Section~\ref{sec:data_compilation}, Table~\ref{tab:examples}), leading to \textit{contradicting} supervision.\footnote{By \textit{contradicting} supervision we mean similar training query-document pairs with opposite labels.} As we hypothesized (Section~\ref{sec:reranking}), re-rankers over-utilize the pre-fetcher score when calculating document relevance, as a defense mechanism (bias) against contradicting supervision, which eventually leads to the degeneration of the re-ranker's term matching mechanism. Inspecting the corresponding weights of the models, we observe that indeed $w_p >> w_s$ across all methods. This effect seems more intense in \bert{-based} re-rankers (\bertlmtc + \drmm or \pacrr), especially those that fine-tune \bertlmtc, possibly because these models perform term matching considering sub-word units, instead of full words. In other words, relying on the neural relevance score ($s_r$) is catastrophic. Similar observations can be made for \ukeu. In both datasets all methods have a large performance gap compared to the \oracle, indicating that there is still large room for improvement, possibly utilizing information beyond text.\vspace{1mm}

\begin{figure}[!ht]
    \centering
    \includegraphics[width=\columnwidth]{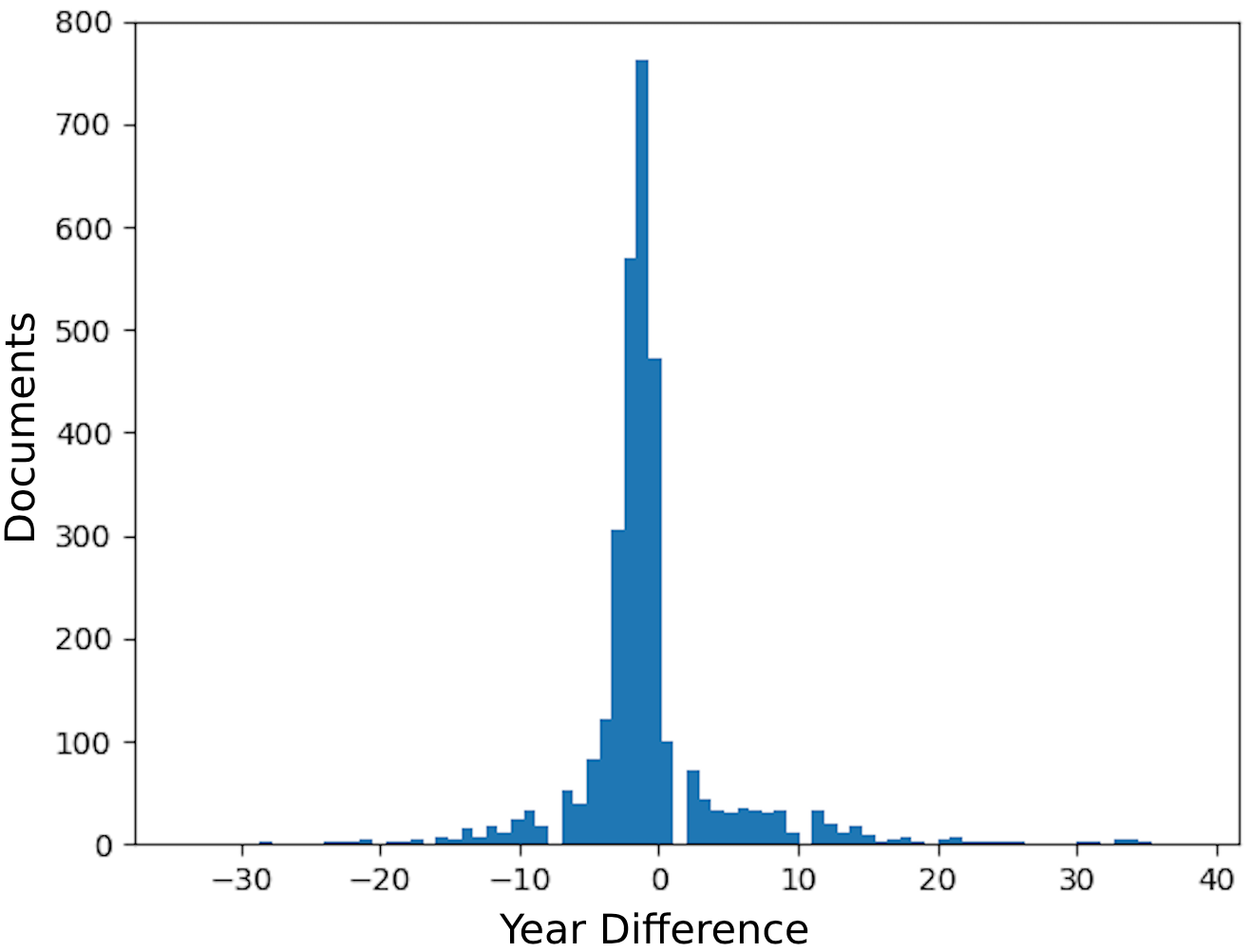}
    \caption{Relevant documents according to their chronological difference with the query on \euuk development data.}
    \label{fig:time}
    \vspace{-2mm}
\end{figure}

\noindent\textbf{Filtering by year:} We have already highlighted the difficulties imposed to our datasets by the frequently amended \eu directives (Section~\ref{sec:data_compilation}, Table~\ref{tab:examples}). Also, recall that each \eu directive defines a deadline (typically 2 years) for the transposition to take place. On the other hand, as we observe in Figure~\ref{fig:time}, \eu directives may already be transposed by earlier legislative acts of member states (the member states act in a proactive manner), or they may delay the transposition for political reasons. In effect, the relevance of a document to a query depends both on the textual content and the time the laws were published. Thus, we filter out documents that are outside a predefined distance (in years) from the query in two ways, \emph{pre-filtering} and \emph{post-filtering}. Pre-filtering is applied to the pre-fetcher, i.e., prior to re-ranking, while post-filtering is applied after the re-ranking. Note that our main goal is to improve re-ranking. We thus apply the filtering scheme to the \ensemble, \drmm and \pacrr. The lower zone of Table~\ref{tab:re-ranking} shows the results of the whole process. In \euuk, the hardest out of the two datasets, the time filtering has a positive impact, improving the results by a large margin. On the other hand, filtering seems to have a minor effect in \ukeu.

\subsection{\euuk $\neq$ \ukeu}

Across experiments, we observe that best practices vary between the \euuk and \ukeu datasets. \euuk benefits from \bertlmtc representations, while in \ukeu context-unaware and domain-agnostic \bm has comparable or better performance than \bertlmtc. Similarly, we observe that time filtering further improves the performance in \euuk, while we have a contradicting effect in \ukeu. Given the overall results, we conclude the two datasets have quite different characteristics. Thus, it is important to consider both \euuk and \ukeu independently, although one may initially consider them to be symmetric.

\section{Related work}
\ir in the legal domain is widely connected with the Competition on Legal Information Extraction/Entailment (\coliee). From 2015 to 2017 \cite{coliee2015,coliee2016,coliee2017}, the task was to retrieve Japanese Civil Code articles given a question, while in \coliee 2018 and 2019 \cite{coliee2018,coliee2019}, the task was to retrieve supporting cases given a short description of an unseen case. However, the texts of these competitions are small compared to our datasets. Also, most submitted systems do not consider recent advances in \ir, i.e, neural ranking models \cite{guo2016,hui2017pacrr,mcdonald2018,cedr2019}, which have recently managed to improve rankings of conventional \ir, or end-to-end neural models which have recently been proposed \cite{Fan2018,khattab2020colbert}. Again, these end-to-end methods were applied on small texts. On the other hand, there has been some work trying to cope with larger queries, i.e., \emph{verbose} or expanded queries, \citep{Paik2014,Gupta2015,cummins2016}. Nonetheless, the considered queries are at most 60 tokens long, contrary to our datasets where, depending on the setting, the average query length is 1.8K or 2.6K tokens (Table~\ref{tab:ir_datasets_stats}). Neural methods greatly rely on text representations, thus \newcite{reimers2019sentence} proposed \sentbert which is trained to compare texts for an \nli task and could thus be used to extract representations suitable for \ir. Towards the same direction, \newcite{Chang2020Pre-training} experimented with several auxiliary tasks to extract better representations. However, the latter two methods have been evaluated on datasets with much smaller texts than the ones we consider.

\section{Conclusions and future work}\vspace{-1mm}
We proposed \doctodoc \ir, a new family of \ir tasks, where the query is an entire document, thus being more challenging than traditional \ir. This family of tasks is particularly useful in regulatory compliance, where organizations need to ensure that their controls comply with the existing legislation. In the absence of publicly available \doctodoc datasets, we compile and release two datasets, containing \eu directives and \uk laws transposing these directives. Experimenting with conventional (\bm) and neural pre-fetchers we showed that a \bert model fine-tuned on an in-domain classification task, i.e., predict \eurovoc concepts, is by far the best pre-fetcher in our datasets. We also showed that neural re-rankers fail to improve the performance, as their term matching mechanisms degenerates, and over-utilize the pre-fetcher score. In the future, we would like to investigate alternatives in exploiting additional information that may be critical in the newly introduced tasks (\euuk, \ukeu). In this direction naively utilizing chronological information leads to vast performance improvement in \euuk dataset. One possible direction is to model the cross-document relations (e.g., amendments) using Graph Convolutional Networks \cite{kipf2016}, while better modeling the dimension of time (i.e., chronological difference between a query and a document) is also crucial. Further on, to better deal with long documents, we plan to investigate text summarization by employing a state-of-the-art neural summarizer, e.g., \bart of \newcite{lewis2020}, or sentence selection techniques, e.g., rationale extraction \cite{lei2016,chang2019}, to find the most important sections or sentences and create shorter and more informative versions of queries/documents.

\bibliography{eacl2021}
\bibliographystyle{acl_natbib}

\appendix

\section{Dataset Compilation: Technical Details}
\label{sec:technical}

In this section, we present the technical details associated with the compilation of both datasets described in the main paper. More specifically we present the procedure of creating both corpora as well as modelling the transposition relations between \eu and \uk entries.  

\subsection{\eu corpus}

The compilation of the \eu corpus is more straightforward than its \uk counterpart but involves some in-domain knowledge to filter unwanted legislation.

\begin{itemize}
    \itemsep0em
    \item We initially download the core metadata associated with each document in the \eu corpus by utilizing the \textsc{sparql} endpoint of the \eu Publications Office (\url{http://publications.europa.eu/webapi/rdf/sparql}) and the \eurlex platform (\url{https://eur-lex.europa.eu}), as a \textsc{rest}{-ful} \textsc{api}.
    \item Following the metadata collection, we proceed to filter out documents based on their type in order to retain only \eu directives and regulations. This involves excluding corrigendums. Corrigendums introduce corrections to prior \eu legislation. Usually these corrections are minimal and change single phrases such as ("In Regulation X, for: ‘… 4 July 2019 …’, read: ‘… 4 July 2015 …’."). Thus these documents lack the context to be both classified and correlated with other documents. \footnote{See \url{https://eur-lex.europa.eu/legal-content/EN/TXT/?qid=1593684165879&uri=CELEX:32004L0038R(02)} as an example.} and decisions, both of which are irrelevant to our use case. The final \eu corpus contains approximately 60k entries.
\end{itemize}

\subsection{\uk corpus}

Compiling the \uk corpus is not as trivial, since the \url{legislation.gov.uk} \textsc{api} is not as evolved and we therefore have to manually crawl large parts of the database to build our corpus.

\begin{itemize}
    \itemsep0em
    \item The collected \uk laws from the \url{legislation.gov.uk} portal form the initial corpus which includes approximately 100k documents. 
    \item Similarly to our processing of the \eu corpus, we only retain documents in specific legislation types (\uk Public General Acts, \uk Local Acts, \uk Statutory Instruments and \uk Ministerial Acts). We then eliminate laws that aim to align English legislation with the rest of the United Kingdom's, more specifically Scotland, Northern Ireland and Wales. The final \uk corpus includes 52K \uk entries.
\end{itemize}

\begin{figure*}[!ht]
    \centering
    \includegraphics[width=\textwidth]{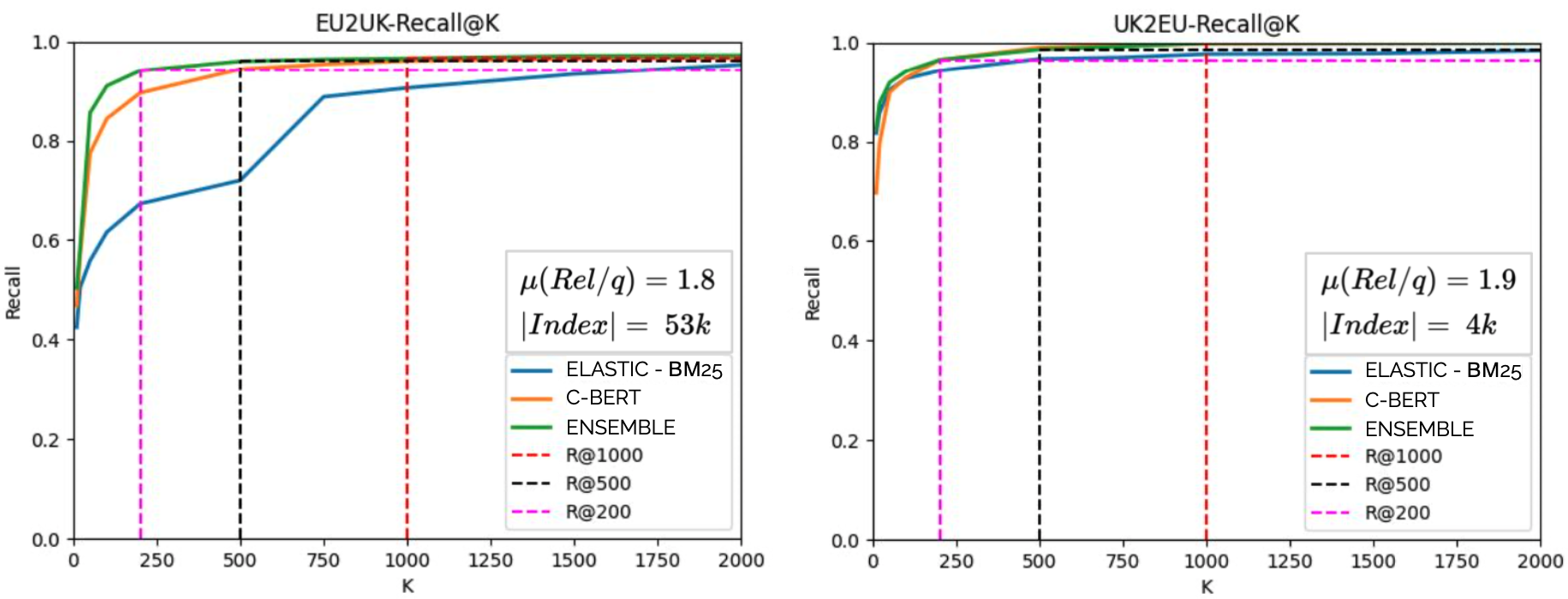}
    \caption{$\mathrm{Recall@k}$, where $k \in [0,2000]$, across the three best pre-fetchers (i.e., \bm, \bertlmtc and \ensemble) on the development dataset.}
    \label{fig:r_k}
\end{figure*}

\subsection{\euuk Transpositions}
\label{sec:appendix_transpositions}
Transpositions are relations between entries in the \eu and \uk corpora which we use to define relevance for our retrieval tasks. Processing these relations is the most challenging aspect of compiling our datasets and involves several steps. 
    
\begin{itemize}
    \itemsep0em
    \item We use the aforementioned \textsc{sparql} endpoint, to retrieve the transpositions between \eu directives and the corresponding \uk regulations that implement them. We initially collect approximately 10k \euuk pairs. In these pairs the transposed \eu law is referred to by its unique portal \textsc{id} but the transposing \uk law is referred to by its title. This is the primary challenge in modelling the transposition relations, since mapping legislation titles to unique entries in our \uk corpus is not trivial. We hypothesize that these relations are manually inserted in the database and therefore human errors make performing exact matches often impossible. Apart from the matching difficulties, some of the pairs in the pool are inserted mistakenly and hence need to be filtered.
    \item We first filter the noisy pairs. Pairs are considered noisy either because they are duplicates or because the do not meet some manually set criteria. In turn, duplication can occur either because identical pairs are inserted more than once or because pairs in which the \uk title is mildly paraphrased are erroneously considered different. Our pool is reduced to 8k pairs after resolving the former and to 7k pairs after also resolving the latter. We further reduce the pool size by filtering pairs in which the \uk title refers to non-English legislation (Scotland, Northern Ireland, Wales or Gibraltar) Non-English legislation usually has an almost identical counterpart within the pure english corpus. \footnote{See \url{https://www.legislation.gov.uk/uksi/2017/407/contents} and \url{https://www.legislation.gov.uk/nisr/2017/81/contents}}. or in which the title does not contain certain keywords (e.g., Act, Regulation, Order, Rule). Documents that do not contain any of these keywords are not officially published in the \url{legislation.gov.uk} portal. Most of these are official releases from national governmental bodies, e.g. Ministries. For instance the \textit{First Annual Report of the Inter-Departmental Ministerial Group on Human Trafficking} is not part of the \uk's national legislation..  
    \item To resolve the matching challenge, we employ a complex matching scheme where for each pair we gradually normalize the \uk title until we find either a singular match or multiple ones. In the latter case, we resolve the matches with heuristics. Our normalizations include lower-casing, leading and trailing phrase removal, punctuation elimination, date removal and manually inserted substitutions. 
    \item After reducing our pair pool and then implementing our matching scheme we can with high confidence present 4k transposition pairs which we use in our datasets.
\end{itemize}

\section{\bert models}
\label{sec:bertmodels}

All \bert variants (\bert, \sentbert, \textsc{legal-bert}) are publicly available from Hugging Face:

\begin{itemize}
    \item \textbf{\bert}: The original \bert pre-trained for Masked Language Modeling (MLM) and Next Sentence Prediction (NSP) in English Wikipedia and Books corpus. Available at \url{https://huggingface.co/nlpaueb/bert-base-uncased-eurlex}.
    \item \textbf{\sentbert}: This is the original \bert fine-tuned in STS-B NLI dataset. Available at  \url{https://huggingface.co/deepset/sentence_bert}.
    \item \textbf{\textsc{legal-bert} (\eurlex)}: This is the original \bert further pre-trained in \eu legislaiton. Available at  \url{https://huggingface.co/nlpaueb/bert-base-uncased-eurlex}.
\end{itemize}

\section{Selecting $\mathrm{k}$ for pre-fetching}

In Section 4.1, we stated that we report $\mathrm{R@k}$ with $\mathrm{k=100}$ in order to evaluate and compare pre-fetching methods. In Figure~\ref{fig:r_k}, we present the performance of the best pre-fetching methods (i.e., \bm, \bertlmtc and \ensemble) for different values of $k \in [0,2000]$ on the development set. We observe that after $k=100$, the \ensemble pre-fetcher has not significant gains in performance, thus we select $k=100$, as a reasonable threshold.

\end{document}